\documentclass[10pt]{article}
\usepackage[preprint]{tmlr}
\usepackage{amsmath,amsfonts,graphicx}
\usepackage{hyperref}
\hypersetup{hidelinks,pdfauthor={Zhenisbek Assylbekov},pdftitle={Two Counterexamples for e-means}}
\title{Two Counterexamples for e-means}
\author{%
\name Zhenisbek Assylbekov \email zassylbe@pfw.edu\\
\addr Department of Mathematical Sciences\\
Purdue University Fort Wayne\\
Fort Wayne, Indiana, USA
}
\date{}
\newtheorem{proposition}{Proposition}
\newtheorem{corollary}{Corollary}

\begin{document}
\maketitle

\begin{abstract}
e-means (Bessa et al., 2026) combines exponential averaging and cluster revival. We give two counterexamples to its published recipe with higher final loss than standard k-means (Lloyd, 1982) from identical initial centers. Mass-based revival can destroy a global optimum and make the ratio of expected losses unbounded even with k-means++ seeding.
\end{abstract}

\section{Introduction}

Standard k-means alternates between assigning the full dataset to its nearest centers and recomputing exact cluster means \citep{lloyd}. These repeated passes make clustering large collections of embeddings costly. \citet{bessa} propose e-means, which updates centers by mini-batch exponential averaging and revives centers with low estimated usage. On a wide array of embedding benchmarks, they report losses below those of 25-epoch Lloyd after 1.4--2.6 epochs.

Both modifications change which observations determine the centers. An observation absent from a batch can affect the optimal partition, while a small, distant cluster can be costly to merge.

Our main construction shows that mass-based revival can destroy a globally optimal solution. For the published parameter laws, e-means' loss ratio to Lloyd grows without bound, including the ratio of expected losses under shared k-means++ seeding (Corollary~\ref{cor:seeding}). An eight-point example then illustrates how forgetting changes a previously observed point's assignment and leads to a worse Lloyd fixed point. Its exact conditional failure probability is certified in Appendix~\ref{app:certificate}.

Our contribution is an explicit failure analysis of the published e-means rule: the revival family extends the rare-cluster failure reported by \citet{faiss5907} with an unbounded loss separation. Related Faiss proposals introduce separation-based and cost-based revival \citep{faiss5908,faiss5915}; our results concern the original mass-based rule.

\citet[Section~3.3]{bessa} already acknowledge limitations of their scaling laws on tiny datasets and data of low intrinsic dimension. Our constructions have intrinsic dimension one. They identify worst-case failure mechanisms within that regime; they make no claim about failure frequencies on embedding benchmarks.

Let \(X=(x_1,\ldots,x_n)\) be a dataset of \(n\) observations in \(\mathbb{R}^d\), allowing repeated values. Let \(C=(c_1,\ldots,c_k)\) denote \(k\) cluster centers in the same space. We measure the training sum of squared distances to the nearest center,
\[
    \Phi(C)=\sum_{i=1}^n\min_{1\le j\le k}\|x_i-c_j\|^2.
\]
Write \(\Phi_\star=\min_C\Phi(C)\) for the global optimum. Throughout, we analyze the printed recipe of \citet[Algorithm~1, Eqs.~5--7, Table~2]{bessa}: exact nearest-center assignments, the \emph{knee at \(0.2\%\)} batch law, initial counts \(N_j=10^{-10}\) and sums \(S_j=N_jc_j\), revival every five batches at threshold \(0.1/k\) of total count, and last-epoch mean finalization. We specify initial centers below; Corollary~\ref{cor:seeding} replaces random-row initialization with shared k-means++ seeding. Section~\ref{sec:revival} distinguishes the printed settings from the pinned library defaults used for validation.

\section{Revival removes the center of a rare cluster}\label{sec:revival}

For e-means, let \(n_j\) and \(s_j\) be the count and coordinate sum assigned to cluster \(j\) in a batch. Its stored count \(N_j\) and sum \(S_j\) are updated as
\[
    N_j\leftarrow(1-\alpha_t)N_j+\alpha_t n_j,\qquad
    S_j\leftarrow(1-\alpha_t)S_j+\alpha_t s_j,\qquad
    c_j=S_j/N_j.
\]
Old assignments lose weight at each update.

Put ten observations at \(1\), ten at \(3\), and one at \(14\). Start both algorithms at \((1,14)\). Lloyd reaches the optimal centers \((2,14)\) in one full pass and stays there. Its squared error is \(10(1-2)^2+10(3-2)^2=20\), and the point at \(14\) has zero error.

We set \(T=210\) epochs so that the printed parameter laws select full batches and a first step of size one, as verified below. e-means initially reaches the same optimal centers. Yet at batch~5 it declares the center at \(14\) dead. Its share \(N_j/\sum_\ell N_\ell=1/21\) is only \(4.76\%\), below the revival threshold \(0.1/k=5\%\) for \(k=2\) \citep[Appendix~H]{bessa}. The rule therefore destroys a globally optimal solution.

\begin{figure}[htbp]
\centering
\includegraphics[width=\linewidth]{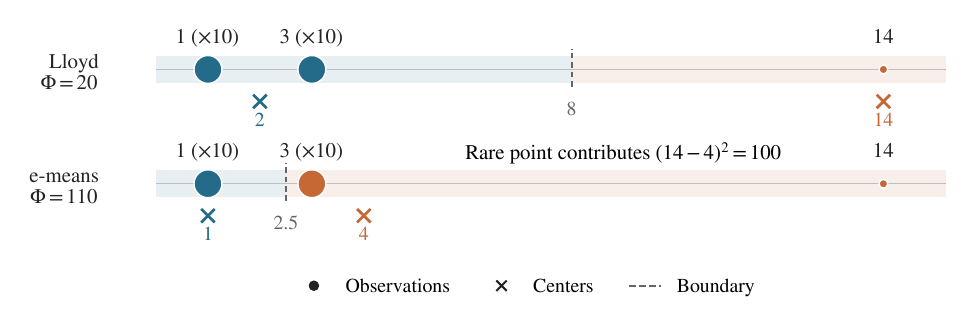}
\caption{Revival discards the isolated point's center. The final centers are \((1,4)\), and the point at \(14\) contributes \(100\) of the total loss \(110\). Lloyd keeps its own center at \(14\) and has loss \(20\).}
\label{fig:revival}
\end{figure}

Faiss splits a live donor to replace the center at \(14\). Its only donor is the center at \(2\). The two new centers lie just to either side of \(2\), so the ones go left and the threes and \(14\) go right. Their final means are \((1,4)\), giving squared error \(110\) (Figure~\ref{fig:revival}). Disabling revival retains Lloyd's loss \(20\).

\paragraph{How large can the loss be?}
Replace ten by an integer \(m\ge10\) and \(14\) by \(D=m+4\).

\begin{proposition}\label{prop:revival}
Put \(m\) observations at each of \(1\) and \(3\), and one at \(D=m+4\). Let \(n=2m+1\), \(k=2\), and \(T=10n\). From the shared initial centers \((1,D)\), the printed recipe returns \((1,4)\), while Lloyd reaches the global optimum \((2,D)\) in one pass. The loss ratio is \((m+1)/2\), for any donor-split perturbation \(0<\delta<1/2\).
\end{proposition}

\paragraph{Proof.}
We choose \(T=10n\) to make the printed laws select full batches and a unit first step. Specifically, the unrounded batch law and the unclipped peak-step law at \(B=n\) satisfy
\[
\begin{aligned}
 \frac{B_{\rm raw}}{n}
   &=0.035\cdot2^{0.13}\cdot10^{0.81}n^{0.69}>2,\\
 \widetilde\alpha_0
   &=0.824\cdot2^{-0.46}\cdot10^{-0.11}n^{0.34}>1.
\end{aligned}
\]
At \(n=21\), the values exceed \(2.0208\) and \(1.3092\), respectively, and both increase with \(n\). Rounding \(B_{\rm raw}\) to the nearest power of two and capping at \(n\) therefore gives \(B=n\); clipping the peak gives \(\alpha_0=1\). All later cosine-scheduled steps lie in \((0,1)\). The first update reaches \((2,D)\), with counts \((2m,1)\). This state persists until the first revival check, which removes the singleton's center because \(1/(2m+1)<0.05\).

The donor splits into \(2(1-\delta)\) and \(2(1+\delta)\), following \citet[Eq.~8]{bessa}. The ones go left, and the threes and \(D\) go right. The left center then remains in \([1,2)\), and the right in \((2,4]\), moving toward \((3m+D)/(m+1)=4\). Their midpoint stays strictly between \(1\) and \(3\), so these assignments persist.

Both children receive half the donor's count, hence start at \(m\). Subsequent counts remain in \([m,m+1]\), so each share is at least \(m/(2m+1)>0.05\). Revival never triggers again. Final-epoch averaging therefore returns \((1,4)\).

The two cuts between distinct data values have costs
\[
\begin{aligned}
  \Phi_{\rm Lloyd}=\Phi_\star&=2m,\\
  \Phi_{e\text{-means}}&=m(3-4)^2+(D-4)^2=m(m+1),\\
  \frac{\Phi_{e\text{-means}}}{\Phi_{\rm Lloyd}}&=\frac{m+1}{2}.
\end{aligned}
\]
These costs establish global optimality of \((2,D)\) and the claimed ratio. The returned centers \((1,4)\) have midpoint \(5/2\), making them a Lloyd fixed point as well. Thus the ratio grows without bound even though every batch contains the rare point.

\paragraph{Shared k-means++ seeding.}
The separation also holds in expectation over a standard randomized initialization.

\begin{corollary}\label{cor:seeding}
For the family and settings of Proposition~\ref{prop:revival}, give both algorithms the same initial centers drawn by standard k-means++ seeding. Then
\[
 \mathbb E[\Phi_{e\text{-means}}]=m(m+1),\qquad
 \mathbb E[\Phi_{\rm Lloyd}]<6m,\qquad
 \frac{\mathbb E[\Phi_{e\text{-means}}]}{\mathbb E[\Phi_{\rm Lloyd}]}
 >\frac{m+1}{6}.
\]
The expectations are over seeding, and the ratio is unbounded already for \(k=2\).
\end{corollary}

\paragraph{Proof.}
k-means++ selects its first center uniformly at random from the observations and the second in proportion to squared distance from it \citep{arthur2007}. Repeated values have zero distance to an existing center, so the possible seed pairs, up to ordering, are exactly those in Table~\ref{tab:seeds}.

\begin{table}[htbp]
\centering
\caption{Final centers for every possible k-means++ seed pair in the revival family.}
\label{tab:seeds}
\begin{tabular}{ccc}
\hline
Initial centers & e-means & Lloyd\\
\hline
\((1,3)\) & \((1,4)\) & \((1,4)\)\\
\((1,D)\) & \((1,4)\) & \((2,D)\)\\
\((3,D)\) & \((1,4)\) & \((2,D)\)\\
\hline
\end{tabular}
\end{table}

For \((1,3)\), the first full-batch update reaches \((1,4)\), with counts \((m,m+1)\); the assignments persist and revival never triggers. For either pair containing \(D\), the first update reaches \((2,D)\), followed by the revival failure proved above. Thus e-means always returns \((1,4)\). With revival disabled, full-batch e-means agrees with Lloyd for every seed pair in Table~\ref{tab:seeds}. Thus revival causes the expected-loss separation.

Lloyd returns the worse solution only when the seed pair is \((1,3)\). This event has probability
\[
 q_m=\frac{4m^2}{2m+1}
 \left[\frac{1}{(m+3)^2+4m}+\frac{1}{(m+1)^2+4m}\right]
 <\frac{4}{m}.
\]
The two terms correspond to the first seed being \(1\) and \(3\). Each denominator exceeds \(m^2\), giving the bound. Consequently,
\[
 \mathbb E[\Phi_{\rm Lloyd}]
 =(1-q_m)2m+q_m m(m+1)
 =2m+q_m m(m-1)<6m.
\]
Since \(\Phi_\star=2m\), Lloyd's expected approximation ratio is below \(3\), while e-means' ratio is \((m+1)/2\). The mass-based revival rule can therefore forfeit the expected approximation guarantee supplied by k-means++ seeding.


\paragraph{Implementation.}
We validated all three seed pairs for \(m=10,20,50\) using \texttt{NumpyEmeans} at \href{https://github.com/facebookresearch/faiss/tree/2b648f0605f2ef3d14c4705350f8a0b214463878/contrib/e_means}{Faiss revision \texttt{2b648f0}}, with explicit \(T=10n,B=n,\alpha_0=1\), and enumerated seeding probabilities exactly. Its automatic recipe gives \(B=4,\alpha_0\approx0.3013\) at \(n=21,T=210,d=3\). Its initial count \(2^{-16}\) is erased by the unit first step. Because revival is disabled by default for \(d\le2\), the checks use \(x\mapsto(x,0,0)\), preserving distances, donor splits, and intrinsic dimension one.

\paragraph{Code availability.}
Standalone code reproducing both counterexamples, the exact k-means++ expectations, and the interval certificate is available at \url{https://github.com/zh3nis/e-means-counterexamples/}.

\section{Forgetting moves a boundary past an absent point}\label{sec:forgetting}

Use two clusters and start both algorithms at
\[
    X=(0,0,15,29,29,29,29,29), \qquad (c_L,c_R)=(15,29).
\]
The initial midpoint is \(22\), so \(15\) joins the two zeros. Lloyd reaches \((5,29)\) in one pass and stays there, with boundary \(17\) and squared error \(2(0-5)^2+(15-5)^2=150\). This is the optimum. The other cut puts \(15\) with the five copies of \(29\), at a cost of \(\tfrac56(29-15)^2=490/3\).

We use 25 epochs, each a full pass through the data. The batch law gives two batches of four per epoch, hence 50 updates. The published peak-step law and cosine schedule give
\[
\begin{aligned}
    \alpha_0&=0.824\cdot2^{-0.46}\cdot4^{0.90}\cdot8^{-0.45}\cdot25^{-0.11}=0.574306\ldots,\\
    \alpha_t&=\alpha_0\bigl(0.55+0.45\cos(\pi t/50)\bigr),\qquad t=0,\ldots,49.
\end{aligned}
\]
\begin{proposition}\label{prop:forgetting}
For these initial centers and settings, independent uniform epoch shuffles make e-means return \((0,80/3)\) with probability \(24/49\), and \((5,29)\) otherwise. Revival never triggers. Lloyd reaches \((5,29)\) in one pass.
\end{proposition}

\begin{figure}[htbp]
\centering
\includegraphics[width=\linewidth]{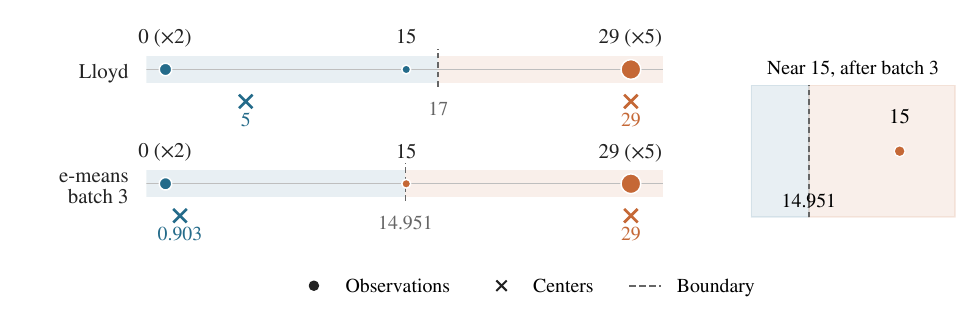}
\caption{Forgetting moves the boundary past \(15\). Colors show nearest-center assignments. Center markers are offset below the line. Parentheses give multiplicities. The inset resolves the crossing.}
\label{fig:forgetting}
\end{figure}

Let the first three batches be \((15,29,29,29)\), \((0,0,29,29)\), and \((0,0,29,29)\). The right center stays at \(29\), while the left moves \(15\to2.637\to0.903\). The crossing threshold is \(c_L=1\), since \((1+29)/2=15\). Batch~3 contains no \(15\), yet moves the boundary to \(14.951<15\) (Figure~\ref{fig:forgetting}). When \(15\) returns, it goes right.

\paragraph{Why the error persists.}
Once \(0\le c_L<1\) and \(15\le c_R\le29\), only zeros go left. Every update preserves these bounds and keeps the midpoint below \(15\). At the end of training, e-means returns ordinary means of the assignments made during the final epoch \citep[Algorithm~1]{bessa}. These are \((0,80/3)\), which retain the worse partition. Thus after 25 passes,
\[
    \Phi_{e\text{-means}}=\frac{490}{3}\approx163.33,
    \qquad \Phi_{\rm Lloyd}=150,
\]
an \(8.89\%\) increase over Lloyd's one-pass loss. The returned centers have midpoint \(40/3<15\), so they are also a Lloyd fixed point.

\paragraph{How often?}
Independently reshuffle the eight indexed observations each epoch. There are \(\binom84=70\) possible first batches. Two disjoint events lead to the worse partition.

If batch~1 omits \(15\) and contains a zero, the boundary crosses \(15\) immediately. There are \(\binom74-\binom54=30\) such batches. Alternatively, batch~1 contains \(15\) and at most one zero, which occurs in \(\binom53+2\binom52=30\) batches. In that case the boundary first crosses at batch~3 exactly when that batch contains both zeros and omits \(15\), as in the example above. There are \(\binom52=10\) such first batches of epoch~2. The failure probability, conditional on the initial centers, is therefore
\[
    \underbrace{\frac{30}{70}}_{3/7\text{: first batch}}+
    \underbrace{\frac{30}{70}\frac{10}{70}}_{3/49\text{: third batch}}
    =\frac{24}{49}.
\]
The first term reflects sensitivity to the first batch when the initial count is almost zero. The second is a later crossing after \(15\) has already been observed, isolating the forgetting mechanism illustrated above. Choosing two distinct data rows uniformly yields the prescribed pair of center values with probability \(5/28\).

Appendix~\ref{app:certificate} completes the proof of the exact probability by excluding later first crossings and revival.

All assignment and revival comparisons in this finite example are strict. Sufficiently small perturbations of the indexed observations and corresponding initial seed rows preserve its partition trajectories, with continuously varying losses.

\section{Conclusion}

Mass-based revival can destroy a global optimum and make e-means finish at a worse Lloyd fixed point. Under shared k-means++ seeding, the ratio of expected losses to Lloyd is still unbounded. A separate example isolates a later assignment change caused by forgetting. Both failures persist under small perturbations of the prescribed data and seeds. The constructions provide explicit tests of the published recipe in its acknowledged low-dimensional regime, and show why small usage alone cannot justify removing a center.

\bibliography{references}
\bibliographystyle{tmlr}

\appendix
\section{Certificate for Proposition~\ref{prop:forgetting}}\label{app:certificate}

The exact probability uses a finite computation with outward-rounded interval arithmetic at 45 decimal digits, interpreting the printed coefficients as exact decimal constants. Here we specify the bounds being certified. Number updates by \(t=1,\ldots,50\), with \(a_t=\alpha_{t-1}\) and \(\eta=10^{-10}\). The retained initial mass and batch weights after update \(t\) are
\[
 p_t=\eta\prod_{s=1}^{t}(1-a_s),\qquad
 w_{t,r}=a_r\prod_{s=r+1}^{t}(1-a_s),\quad 1\le r\le t.
\]
Let \(u_r\in\{0,1\}\) indicate whether batch \(r\) contains \(15\), and let \(z_r\in\{0,1,2\}\) count its zeros. Before a first crossing, the right center is \(29\), and the left center satisfies
\[
 \frac{c_L^{(t)}}{15}
 =\frac{p_t+\sum_{r=1}^t w_{t,r}u_r}
 {p_t+\sum_{r=1}^t w_{t,r}(u_r+z_r)}.
\]
Each complete epoch contributes one \(15\) and two zeros. Pair the weights by epoch, assigning weight zero to an unseen second batch, and write
\[
 \begin{aligned}
 A_t&=p_t+\sum_{(p,q)}\min(p,q),\\
 Z_t^-&=2\sum_{(p,q)}\min(p,q),\qquad
 Z_t^+=2\sum_{(p,q)}\max(p,q).
 \end{aligned}
\]
Putting \(15\) in the lighter batch and both zeros in the heavier bounds any possible first crossing by
\[
 c_L^{(t)}\ge\frac{15A_t}{A_t+Z_t^+}>1.0275>1,
 \qquad t=4,\ldots,50.
\]
For the first three updates, direct evaluation for the possible pairs \((u_1,z_1),(u_3,z_3)\), with \((u_2,z_2)=(1-u_1,2-z_1)\), gives exactly the two crossing events counted in Section~\ref{sec:forgetting}. Their respective counts among the \(70^2\) equally likely indexed batch pairs are \(2100\) and \(300\); the other \(2500\) have no crossing.

To exclude revival, use a joint induction over the checks. Before a first crossing, updates keep \(0\le c_L\le15\) and \(c_R=29\); after a crossing, the invariant in Section~\ref{sec:forgetting} gives \(0\le c_L<1\) and \(15\le c_R\le29\). In both cases zeros go left and copies of \(29\) go right. Assuming no earlier revival, the weight formulas therefore imply at each check
\[
 \frac{N_L}{N_L+N_R}
 \ge\frac{p_t+Z_t^-}{2p_t+4\sum_{r=1}^t w_{t,r}}
 >0.06410>0.05,
 \qquad t=5,10,\ldots,50.
\]
Every four-row batch contains at least one \(29\), so the right share is at least \(1/4\). Neither center is revived. This preserves the geometric conditions until the next check and closes the induction. The invariance argument in Section~\ref{sec:forgetting} then determines the final partition from whether a crossing occurred.

The repository's \texttt{counterexamples.py} verifies these bounds and the early-event classification using \texttt{mpmath.iv}, counts the \(4900\) indexed batch pairs, and reproduces the displayed center trace and losses against Lloyd.
\end{document}